\documentclass{article}
\pdfoutput=1
\usepackage[T1]{fontenc}
\usepackage[utf8]{inputenc}
\usepackage[english]{babel}
\usepackage{amsthm}
\usepackage{newtxtext,newtxmath}
\usepackage[disable=footnote]{microtype}

\usepackage[letterpaper,margin=1in]{geometry}

\usepackage{amsmath,amsthm} 
\usepackage{mathtools}
\usepackage{graphicx}
\usepackage[dvipsnames]{xcolor}
\usepackage[colorlinks=true,linkcolor=blue,citecolor=blue,urlcolor=blue]{hyperref}
\usepackage{url}
\usepackage{booktabs}
\usepackage{array,multirow}
\usepackage[numbers,sort&compress]{natbib}
\usepackage{caption}
\usepackage{verbatim}
\usepackage{setspace}
\usepackage{enumitem}
\setlist{nosep,leftmargin=*}

\theoremstyle{remark}

\usepackage{titlesec}
\titleformat{\section}{\large\bfseries}{\thesection}{1em}{}
\titlespacing{\section}{0pt}{8pt plus 2pt}{4pt plus 2pt}
\titleformat{\subsection}{\normalsize\bfseries}{\thesubsection}{1em}{}
\titlespacing{\subsection}{0pt}{6pt plus 2pt}{3pt plus 2pt}
\titleformat{\subsubsection}{\normalsize\bfseries}{\thesubsubsection}{1em}{}
\titlespacing{\subsubsection}{0pt}{5pt plus 2pt}{2pt plus 2pt}

\usepackage{fancyhdr}
\providecommand{\keywords}[1]{\par\addvspace{\medskipamount}
\noindent\textbf{\emph{Keywords:}} #1\par}

\title{Beyond Correctness: \\
Toward Automated Novelty Verification with Lean 4}

\author{
  Ayrton Porto\\
  \small
  Universidad Nacional del Centro de la Provincia de Buenos Aires (UNICEN)\\
  \small\textit{This work was conducted independently.}\\
  \texttt{ayrporto@gmail.com}
}

\date{July 2026}

\begin{document}

\maketitle

\begin{center}
\small
\textbf{Preprint.} Work in progress --- comments welcome.\\
\url{https://github.com/ayrtonporto/avid-journal}
\end{center}

\begin{abstract}
Artificial intelligence systems applied to mathematics verify correctness but not novelty: an automatically generated theorem can compile in Lean without errors and yet be an already known result. This article presents AViD Journal, a pipeline that receives a LaTeX article, formalizes its statements in Lean~4, and issues a novelty verdict through a decision tree over three dimensions: prior existence in a formal corpus (Mathlib) and an informal one (TheoremSearch and Matlas, with temporal filter and LLM judge), non-triviality via automatic tactics, and structural distance between proofs measured as Jaccard distance over premise sets.

Evaluation on papers withdrawn from arXiv due to declared duplication produced a result more informative than any performance measure: the identification of three obstacles that limit the approach regardless of this implementation. First, successful compilation of a Lean file does not guarantee semantic fidelity. Second, the recall ceiling is imposed by the coverage of theorem indices, not by the similarity metric. Third, arXiv removes the source code of articles upon withdrawal, compromising the reproducibility of any benchmark built upon them.
\end{abstract}

\keywords{automated theorem proving, novelty verification, Lean~4, Mathlib, semantic search, proof distance, Jaccard similarity, formal mathematics}
\section{Introduction}

The \textit{First Proof} project~\cite{abouzaid2026}, in which eleven mathematicians contributed unpublished research problems, was designed to separate genuine reasoning from mere literature search. Its second batch, evaluated by human judges~\cite{firstproof2b}, documented a concrete failure mode of language models applied to mathematics: systems performed better when the problem was structurally similar to an already published result, and in several cases solved it by translating a prior proof from the literature, reusing even its terminology and notation line by line, without citing the source, to the point that a human reviewer would have flagged it as plagiarism. What is missing, then, is not the ability to generate correct proofs, but a systematic method to assess whether a proof is genuinely new. Models can generate statements that compile, that are correct, and yet are not new.

The problem is twofold. On the one hand, current mathematical AI systems (Axiom, Harmonic, DeepSeek-Prover) verify correctness but not novelty. On the other, the naive criterion that equates novelty with absence from Mathlib~\cite{kasaura2025} fails in two directions: it marks as new trivial theorems that any automated tactic from 2026 closes without requiring a genuine mathematical idea, and it misses results that are not formalized but are published in the informal literature.

Tao~\cite{tao2025}, in his notes on \textit{Machine-Assisted Proof}, frames the problem in terms of scale: formal verification is human-labor intensive, to the point that today it is not feasible to formalize in real time a significant fraction of research articles, while automated assistance promises mathematical exploration at scales currently unattainable. Tao directs that argument to the control of \emph{correctness} and to the scarcity of reviewers willing to verify long proofs line by line. The same argument, in our view, transfers to the control of \emph{novelty}: if theorem generation accelerates, manually evaluating whether each result is truly new becomes as infeasible as that line-by-line verification, and the party proposing the theorems is itself a model.

The urgency of the problem was illustrated while this work was being written. In July 2026, Levent Alp\"oge, a mathematician at Anthropic, announced an explicit counterexample to the Jacobian conjecture, open since Keller formulated it in 1939, found with the assistance of a language model and so brief that its verification is elementary: a symbolic calculation of seconds, independently formalized in Lean and Isabelle within hours against a pre-registered statement of the conjecture~\cite{alpoge2026}. The counterexample extends immediately to all dimensions $\geq 3$, and within days families of higher-dimensional counterexamples were circulating; the episode is also part of a broader wave of AI-assisted refutations of other open conjectures. Such an episode produces in concentrated form the material that motivates this work: results descending from the same discovery, obtained in parallel by different authors, where the question of whether a statement was already established by another is genuine and costly to answer manually. It is furthermore the regime in which the coverage limitation we document in section~\ref{ssec:cobertura} does not apply: the potentially duplicated works are contemporary and immediately publicly available.

This article presents AViD Journal (Automated Verification in Demonstrations), a system that addresses that question operationally: it receives a LaTeX article, formalizes its statements in Lean~4, and issues a novelty verdict through a three-dimensional decision tree.

\subsection{Contributions}

\begin{enumerate}
\item \textbf{A complete novelty verification pipeline.} The system traverses the entire path, from LaTeX source to verdict: parsing of mathematical environments, formalization in Lean~4 with multi-model abstraction, and an eight-verdict decision tree over three dimensions: prior existence in a formal corpus (Mathlib, via Leandex) and an informal one (TheoremSearch and Matlas, with temporal filter and LLM judge), non-triviality via Lean's automatic tactics, and structural distance between proofs as Jaccard distance over premise sets.

\item \textbf{A benchmark of papers withdrawn for duplication.} A dataset of papers withdrawn from arXiv whose withdrawal comment declares duplication of prior results, with controls matched by category and year, and the analysis of the retrievability of their duplicators.

\item \textbf{A map of the failure modes of the approach.} The central contribution of the evaluation is not a performance metric but the characterization of three walls: (a) that a Lean file compiles does not guarantee that the formalization is faithful to the original statement; (b) the results that duplication rediscovers tend to predate the preprint era, and are therefore invisible to available corpora; and (c) arXiv removes the LaTeX source upon withdrawing a paper, which limits the reproducibility of any benchmark built upon them.
\end{enumerate}

\subsection{Structure of the article}

Section~\ref{sec:related} situates this work with respect to the research fronts that intersect in novelty verification: statement search infrastructure, novelty filters in conjecture generation, proof identity, and the distinction between correctness and novelty. Section~\ref{sec:pipeline} describes the pipeline: parser, formalization, reference corpora, the three dimensions, and the verdict tree. Section~\ref{sec:instrumentos} validates each instrument separately. Section~\ref{sec:experimentos} reports the evaluation on withdrawn papers and characterizes the failure modes found. Section~\ref{sec:conclusion} closes with what was built, the remaining limitations, and the work ahead.

\section{Related Work}

\label{sec:related}

\subsection{Theorem Search Engines and Infrastructure}

\textbf{TheoremSearch}~\cite{theoremsearch2026} indexes 9.2 million statements extracted from arXiv and seven additional sources, represents each theorem with a short natural language description for embedding, and exposes a public API without authentication. Its documented motivation (arXiv withdrawals due to duplication) is the same one that motivates our experiment with withdrawn papers. TheoremSearch finds similar statements; AViD adds the verdict layer.

\textbf{Matlas}~\cite{matlas2026} extracts 8.07 million statements from 435,000 peer-reviewed articles spanning 1826 to 2025, sourced from 180 journals selected by ICM citation criteria, plus 1,900 textbooks. The two sources are complementary in temporal coverage: TheoremSearch covers arXiv since 1991, Matlas covers peer-reviewed journals since 1826. AViD consumes them as the two branches of its informal corpus, and section~\ref{sec:experimentos} reports how far that combined coverage reaches.

The Mathlib search engine space is saturated. \textbf{Leandex}~\cite{leandex2026}, Project Numina's semantic search engine, offers search over Lean~4 declarations; \textbf{Lean Finder}~\cite{leanfinder2025} incorporates user intent into semantic search over Mathlib. AViD does not compete in that space: it consumes one of them as infrastructure (Leandex is the backend for D1~C$_F$) and adds the decision layer that none offers. A recent survey on mathematical AI~\cite{survey2026} cites statement search engines as infrastructure for locating already established theorems, which situates the problem AViD addresses as recognized by the field.

Two recent works are relevant as infrastructure, not as competitors. \textbf{TheoremGraph}~\cite{theoremgraph2026} builds a statement-level dependency graph that unifies the informal (11.7M theorem-type environments from arXiv) and the formal (LeanGraph, 388,105 Lean~4 nodes); we evaluated using its graph dependencies to measure distance between proofs in D3, but ultimately stayed with the direct premise set of each proof. In a complementary direction, \textbf{COMPOSE}~\cite{compose2026} predicts plausible future statements by combining an article's citation graph with Mathlib's formal dependencies, generating conjectures rather than arbitrating their novelty.

\subsection{Novelty Filters in Conjecture Generation}

\textbf{LeanConjecturer}~\cite{leanconjecturer2025} filters novelty with \texttt{exact?} against Mathlib and non-triviality with \texttt{aesop}: it generated 12,289 conjectures from 40 seed files, of which 3,776 proved syntactically valid and non-trivial. These are exactly the mechanisms that AViD implements as D1 (\texttt{exact?} as fallback for C$_F$) and D2 (\texttt{aesop} in \texttt{T\_AUTO}). AViD's contribution is not inventing these filters but composing them into a decision tree that issues a verdict, adding D3 and an informal branch with temporal filter, and evaluating them against external ground truth. Work on pruning redundant conjectures~\cite{mining2024} and synthetic generation via forward reasoning~\cite{synthetic2024} illustrates the same need from the generation side.

\subsection{Proof Identity and Similarity}

The Jaccard distance over premise sets that AViD uses in D3 borrows its representation from the premise selection literature, where a theorem is characterized by the premises of its proof: MaSh~\cite{kuhlwein2013}, \textbf{Kaliszyk and Urban}~\cite{kaliszyk2014} with IDF-weighted $k$-NN, and \textbf{Magnushammer}~\cite{magnushammer2024} with transformers. But the task is the inverse: those systems \textit{select} premises to construct a proof from a statement, whereas D3 uses the set of premises \textit{already employed} as a fingerprint to compare finished proofs. The only directly inherited component is the filtering: the \textit{math filter} of \textbf{Piotrowski et al.}~\cite{piotrowski2023}, which uses Mathlib names as a whitelist to discard technical lemmas, is the same recipe applied by the filters prior to D3's Jaccard.

The Mathlib network analysis by Li et al.~\cite{network2026}, over a graph of 308,129 declarations and 8.4 million edges, shows that the most cited lemmas in the library are technical infrastructure rather than deep mathematics: equality reflexivity (\texttt{Eq.refl}) is the second highest in-degree node, with 69,580 citations, while the Chinese Remainder Theorem does not even appear among the top hundred. That is, centrality in the graph measures how ubiquitous a lemma is, not how much mathematical content it contributes. This justifies D3's Filter~1: before computing the Jaccard between two proofs, AViD discards premises from the \texttt{Init.} and \texttt{Lean.} namespaces, which are precisely those omnipresent technical lemmas. Without the filter, two proofs would appear similar merely for sharing the basic machinery of the language; with it, the Jaccard reflects the premises that truly distinguish mathematical content.

\textbf{Yoo}~\cite{yoo2025} represents theorems as \textit{proof vectors} over axiom systems and compares them with cosine, Euclidean, or Jaccard similarity. It is the closest work in conceptual tool but with opposite purpose: the Atlas organizes theorems by structural similarity; AViD uses that similarity to decide whether a proof is redundant. \textbf{Huch}~\cite{huch2022} documents the scale-free in-degree distribution in the AFP, relevant for future work with IDF-weighted premises.

\subsection{Correctness versus Novelty}

Existing paper-to-Lean pipelines verify that a proof is correct, not that it is new; novelty verification is delegated to human reviewers. AViD is explicitly that delegated layer. A complete automated review system would require both dimensions, and section~\ref{sec:experimentos} shows that the first is not a sufficient condition for the second: a Lean file can compile without errors and yet not be a faithful formalization of the original statement.

\subsection{Positioning}

\begin{table}[htbp]
\centering
\small
\setlength{\tabcolsep}{4pt}
\begin{tabular}{lccccc}
\toprule
System & Existence & Triviality & Proof dist. & Informal corpus & Verdict \\
\midrule
Kasaura et al.~\cite{kasaura2025}      & Mathlib            & --- & --- & --- & yes/no \\
LeanConjecturer~\cite{leanconjecturer2025} & \texttt{exact?} & \texttt{aesop} & --- & --- & --- \\
TheoremSearch~\cite{theoremsearch2026} & 9.2M statements    & --- & --- & \checkmark & --- \\
Matlas~\cite{matlas2026}               & 8.07M statements   & --- & --- & \checkmark & --- \\
Atlas~\cite{yoo2025}                   & ---                & --- & Jaccard & --- & --- \\
AViD (this work)                      & Mathlib + informal & 6 tactics & Jaccard$^{*}$ & TS + Matlas & 8 verdicts \\
\bottomrule
\end{tabular}
\caption{Positioning of AViD with respect to prior work. Prior works build infrastructure pieces; AViD composes them into a pipeline that goes from LaTeX source to verdict. $^{*}$D3 is implemented and validated in isolation (section~\ref{sec:instrumentos}), but is not exercised within the decision tree in the reported experiments.}
\label{tab:posicionamiento}
\end{table}

The difference lies not in any individual cell but in the composition: prior works build infrastructure pieces---search, formalization, similarity---; AViD assembles them into a pipeline that receives a LaTeX paper and returns a verdict.

\section{The AViD Pipeline}

\label{sec:pipeline}

The system receives a \texttt{.tex} file, extracts its mathematical blocks, formalizes them in Lean~4, and applies a three-dimensional decision tree to issue a novelty verdict. This section describes each stage with the detail necessary for its conceptual reproduction.

\subsection{LaTeX Ingestion and Parsing}

The parser extracts mathematical environments from the LaTeX source: \texttt{theorem}, \texttt{lemma}, \texttt{proposition}, \texttt{corollary}, \texttt{definition}, and their Spanish variants (\texttt{teorema}, \texttt{lema}, \texttt{proposición}, \texttt{definición}, \texttt{corolario}). It also recognizes abbreviated variants (\texttt{thm}, \texttt{lem}, \texttt{prop}, \texttt{cor}, \texttt{defn}) and user-defined environments via \verb|\newtheorem| and \verb|\theoremstyle|.

Dependencies between blocks are extracted from cross-references \verb|\ref{}| and \verb|\cite{}|. With them the parser builds a directed acyclic graph and applies a topological ordering (Kahn's algorithm) to determine the formalization sequence: blocks without dependencies are processed first; those that depend on others, afterwards.

Not all detected environments proceed to the next stage. The orchestrator only formalizes types considered \textit{formalizable}: \texttt{theorem}, \texttt{lemma}, \texttt{proposition}, \texttt{corollary}, and their variants. Environments such as \texttt{remark}, \texttt{example}, \texttt{proof}, and \texttt{definition} (when not required as a dependency) are extracted but not formalized.

\textbf{Known limitation.} The parser operates via compiled regular expressions with a fixed list of environment names. Papers using AMS-\TeX{} (\texttt{\textbackslash documentstyle\{amsppt\}}) or defining environments with idiosyncratic names (\verb|\begin{inizio}|, \verb|\begin{numero}|) are not parseable by this version. Of the 33 candidates in the withdrawn dataset, 7 proved non-viable for this reason.

\subsection{Formalization in Lean 4}

\subsubsection{Multi-model Abstraction}

The formalization of each block (statement and proof) is delegated to a language model through an abstract \texttt{ModelProvider} interface. The current implementation supports eight backends: Claude Code (agentic mode, via CLI), Anthropic API, OpenAI, DeepSeek, OpenRouter, Mistral, Gemini, and OpenCode Go. The experiments reported in section~\ref{sec:experimentos} use Qwen 3.7-max through OpenCode Go, selected by empirical comparison across backends.

The architecture distinguishes two families of providers. \textit{Agentic} ones handle their own verification cycle: the model receives the prompt, generates Lean code, attempts to compile, and corrects iteratively. \textit{API} ones use an external verification cycle: the system sends the prompt, receives the response, extracts the Lean code, compiles with \texttt{lake env lean}, and re-sends errors as feedback for the next iteration. This iterative prompting scheme with compiler feedback adopts the strategy of the Numina-Lean-Agent~\cite{leandex2026}.

Each block is classified into one of four modes according to its estimated complexity: \texttt{SIMPLE} (5 maximum correction rounds), \texttt{MEDIUM} (15 rounds), \texttt{HARD} (30 rounds), and \texttt{EXTERNAL} (no model; an axiom with source reference is emitted).

\subsubsection{Shared Lean Project}

All papers share a single Lean~4 project (v4.29.0) with a Mathlib build (8,247 \texttt{.olean} files). Each paper is hosted as a submodule under \texttt{Papers/<Slug>/}. Formalized blocks accumulate incrementally in a \texttt{Paper.lean} file and are registered in an index that the orchestrator consults before searching Mathlib.

The formalization success criterion is demanding: the file must compile without errors, without \texttt{sorry} in any declaration, and must contain at least one substantive declaration (\texttt{theorem}, \texttt{lemma}, \texttt{def}). Empty files or files with only imports are rejected. Section~\ref{sec:experimentos} documents why this criterion, demanding as it is, does not guarantee semantic fidelity.

\subsubsection{Statement-only Mode}

For experiments where only the statement is required, the system operates in \textit{statement-only} mode: the model generates the Lean statement with \texttt{:= by sorry} and the success criterion is relaxed to accept the presence of \texttt{sorry} as long as there are no compilation errors. This mode is not a native capability of the codebase but a configuration of prompts and the acceptance criterion. All experiments on real papers reported in this work operate in this mode, which implies that D3 (which requires premises from a proof) is not exercised in them.

\subsection{Reference Corpora}

The system compares each statement against three corpora.

\textbf{C$_F$ (formal corpus).} Mathlib v4.29.0, accessed through the Leandex API~\cite{leandex2026}, Project Numina's semantic search engine. Leandex indexes Mathlib statements and returns matches with their proof status. The current API version (v2) does not provide similarity scores; the system uses the ordering provided by the API and discards results whose status is \texttt{statement\_only}. The absence of a threshold is a limitation with consequences for the interpretation of D1 C$_F$ results (section~\ref{sec:instrumentos}).

\textbf{C$_I$ (informal corpus).} Two sources complementary in temporal coverage: TheoremSearch~\cite{theoremsearch2026}, with 9.2 million statements extracted from arXiv and seven additional sources, and Matlas~\cite{matlas2026}, with 8.07 million statements from peer-reviewed journals since 1826. Both expose public APIs without authentication. Retrieved candidates are filtered with MiniLM embeddings (cosine similarity threshold $\geq 0.40$) and then pass through the filters described in section~\ref{ssec:d1} before reaching the LLM judge.

\textbf{Own corpus.} As the orchestrator formalizes blocks from a paper, it accumulates them in a local index. Blocks already processed within the same paper are consulted before searching Mathlib, preventing the system from marking as already existing a theorem that the paper itself has just introduced.

\subsection{The Three Dimensions}

\subsubsection{D1: Prior Non-existence}
\label{ssec:d1}

D1 verifies whether the statement already appears in the formal corpus (C$_F$) or the informal corpus (C$_I$). The evaluation follows a fixed order with short-circuiting: if C$_F$ finds a match, C$_I$ is not executed.

\textbf{C$_F$ branch.} The Leandex API is queried with the statement in natural language. If Leandex returns at least one result whose status is not \texttt{statement\_only}, the theorem is considered to exist in Mathlib. No similarity threshold is applied, because the API does not provide scores.

\textbf{\texttt{exact?} fallback.} If Leandex finds no match, the system executes \texttt{example : $\tau$ := by exact?} with a 15-second budget on the Lean environment with full Mathlib. If the tactic finds a theorem that closes the goal, it is treated as a C$_F$ match. This tactic is classified as D1 and not D2 because its function is to find an already proven theorem, not to close the goal by automation.

\textbf{C$_I$ branch.} Activated only if C$_F$ found no match. Stage A queries TheoremSearch and Matlas with the statement, previously cleaned of LaTeX markup. Candidates surviving the embedding threshold pass through two structural filters before stage B, in which an LLM judge (temperature $=0$) compares each candidate with the original statement and issues one of four verdicts: \texttt{equivalent}, \texttt{generalization}, \texttt{specialization}, \texttt{different}.

\textbf{Temporal filter.} A valid duplicator must predate the work it duplicates. The system discards any candidate whose publication date is later than that of the paper under evaluation. For arXiv candidates, the date is derived from the identifier, which encodes year and month in three distinct historical formats (\texttt{YYMM.NNNNN}, \texttt{cat/YYMMNNN}, and the variant without category prefix). For Matlas candidates, identified by DOI rather than arXiv ID, the publication year provided by the source is used. Candidates without a retrievable date are not discarded: they pass to stage B and are tallied separately, so that their presence is recorded rather than producing a silent discard. If the paper under evaluation has no parseable date, the filter is deactivated.

Without this filter, the system can issue a non-novelty verdict against a work \textit{later} than the one evaluated, which is logically impossible as duplication. Section~\ref{sec:experimentos} documents a concrete case and its correction.

\textbf{Self-paper exclusion.} The C$_I$ indices contain the papers the system evaluates. Without explicit exclusion, a paper retrieves itself as a candidate and the judge classifies it as \texttt{equivalent}, producing a duplication false positive. The system passes the evaluated paper's identifier as an exclusion parameter to sources that support it natively, and applies a post-search filter on those that do not.

\subsubsection{D2: Non-triviality}

D2 verifies whether the statement can be proved exclusively with Lean's standard automatic tactics. If any tactic closes \texttt{example : $\tau$ := by T}, the theorem is considered trivial.

\textbf{Tactics, order, and budgets.} The set \texttt{T\_AUTO} is executed in this order: \texttt{decide}, \texttt{norm\_num}, \texttt{simp}, \texttt{omega}, \texttt{tauto}, \texttt{aesop}. The first five have 10 seconds each; \texttt{aesop} has 30. The order prioritizes cheap and specific tactics; \texttt{aesop}, which performs a more exhaustive search, goes last. Execution stops at the first tactic that closes the goal.

Each tactic invocation requires loading the Lean environment with full Mathlib. Without pre-warming, that startup dominates execution time; the times reported in section~\ref{sec:instrumentos} correspond to invocations on an already loaded environment.

\textbf{\texttt{norm\_num} blacklist.} The \texttt{norm\_num} tactic in Mathlib v4.29.0 includes a shortcut that closes \texttt{Irrational (Real.sqrt 2)}. To avoid that false positive, \texttt{norm\_num} is excluded when the statement contains the word \texttt{Irrational}.

\textbf{\texttt{exact?} does not belong to D2.} The tactic was relocated to D1 as C$_F$ fallback for a semantic reason: it searches for an existing theorem in the environment, which is verification of prior existence and not of triviality.

\textbf{Sensitivity to the formalized statement.} The D2 result is not a fixed property of the theorem but of the triple (formalized statement, tactic, budget). The same informal theorem can produce formalizations that \texttt{aesop} closes in seconds or that exceed the budget by an order of magnitude, depending on which definitions the formalizer chose. Section~\ref{sec:instrumentos} documents a concrete case.

\subsubsection{D3: Structural Premise Distance}

D3 measures how different a proof is from another existing in Mathlib. The metric is the Jaccard distance over the sets of premises (lemmas, theorems, definitions) used in each proof.

\textbf{Premise extraction.} Premises are extracted with a standalone Lean tool that traverses the \texttt{InfoTree} and collects, for each \texttt{TermInfo} node, the referenced constant, its definition location, and the module containing it. References occurring at the constant's own definition site are excluded, to prevent a theorem from registering itself as a premise.

\textbf{Canonical identity and deduplication.} Two premises with the same definition location represent the same logical object, even if they appear in different positions of the proof. Before any filtering, premises are deduplicated by this canonical identity.

\textbf{Filtering pipeline.} Before computing Jaccard, premises pass through two filters:

\begin{enumerate}
\item \textbf{Namespace blacklist.} Premises whose module begins with \texttt{Init.} or \texttt{Lean.} are removed. These namespaces contain kernel type constructors, typeclass instances, and tactic internals that the elaborator resolves automatically; they do not reflect mathematical content and would artificially dominate the intersection. The Mathlib network analysis by Li et al.~\cite{network2026} supports this criterion: over a graph of 308,129 declarations and 8.4 million edges, they find that centrality captures language infrastructure more than mathematical relevance.

\item \textbf{Statement premises.} Premises whose position in the source file falls within the statement's line range are removed. Constants appearing in the statement (hypotheses, types, local definitions) are part of the signature and not of the proof; comparing them artificially inflates similarity.
\end{enumerate}

\textbf{Computation.} Let $P(A)$ and $P(B)$ be the post-filter and deduplicated premise sets. The Jaccard distance is
\[
d_J(A,B) \;=\; 1 - \frac{|P(A) \cap P(B)|}{|P(A) \cup P(B)|}.
\]
The decision threshold is $\theta = 0.5$: if the distance exceeds it, the proofs are considered structurally distinct. If either set is empty after filtering, no distance is emitted and the verdict is \texttt{INCONCLUSIVE}.

\textbf{Calibration status.} The threshold $\theta = 0.5$ is the initial design value and has not been calibrated against a broad set of pairs. Section~\ref{sec:instrumentos} reports its behavior on five pairs with human judgment, of which only two prove informative for calibration.

\subsection{Verdict Tree}

The orchestrator traverses the three dimensions in increasing cost order (figure~\ref{fig:arbol}).

\begin{figure}[htbp]
\centering
\small
\begin{verbatim}
D2 (triviality): does any T_AUTO tactic close tau?
  |
  +-- YES -> NOT_NOVEL_trivial                      (END)
  |
  +-- NO  -> D1 C_F (Leandex)
             |
             +-- match    -> D3 (if premises available)
             |            +-- distant         -> PROOF_NOVELTY
             |            +-- close           -> NOT_NOVEL_redundant
             |            +-- empty sets      -> INCONCLUSIVE
             |            +-- not available   -> MATCH_PENDING_D3
             |
             +-- no match -> exact? (fallback)
                              |
                              +-- match -> same path as C_F
                              |
                              +-- no match -> D1 C_I
                                               (TheoremSearch + Matlas,
                                                temporal filter, LLM judge)
                                   +-- equivalent        -> KNOWN_IN_LITERATURE
                                   +-- generalization /
                                       specialization    -> GRAY_ZONE
                                   +-- different or
                                       no candidates     -> STATEMENT_NOVELTY
\end{verbatim}
\caption{Orchestrator decision tree. The dimensions are evaluated in increasing cost order: D2 is local, D1 C$_F$ is an HTTP query, D1 C$_I$ involves several APIs and an LLM judge, and D3 requires premise extraction.}
\label{fig:arbol}
\end{figure}

\textbf{Short-circuit logic.} D2 is evaluated first because it is local and cheap: if the theorem is trivial, the tree terminates without consulting external APIs. D1 C$_F$ is second because it is a fast HTTP query. D1 C$_I$ is third because it involves several APIs and a judge call. D3 is the most expensive and only runs when there is a C$_F$ match that requires deciding whether the proof is novel or redundant.

\textbf{The eight verdicts.}

\begin{enumerate}
\item \texttt{STATEMENT\_NOVELTY}: no match in C$_F$ or C$_I$.
\item \texttt{PROOF\_NOVELTY}: match in C$_F$, D3 indicates distant proofs.
\item \texttt{KNOWN\_IN\_LITERATURE}: match in C$_I$, no match in C$_F$.
\item \texttt{NOT\_NOVEL\_redundant}: match in C$_F$, D3 indicates same proof.
\item \texttt{NOT\_NOVEL\_trivial}: D2 closes with standard tactic.
\item \texttt{GRAY\_ZONE}: the judge classifies the match as generalization or specialization; requires human review.
\item \texttt{MATCH\_PENDING\_D3}: match in C$_F$, D3 not available.
\item \texttt{INCONCLUSIVE}: D3 executed, premise sets empty after filtering.
\end{enumerate}

Three of the eight verdicts (\texttt{PROOF\_NOVELTY}, \texttt{NOT\_NOVEL\_redundant}, \texttt{INCONCLUSIVE}) depend on D3 and, therefore, are not emitted in the reported experiments, which operate in statement-only mode.\footnote{The source code uses Spanish identifiers: \texttt{NOVEDAD\_ENUNCIADO}, \texttt{NOVEDAD\_DEMOSTRACION}, \texttt{NO\_NOVEDOSO\_redundante}, etc. The verdict names are translated throughout this paper for readability; the mapping is one-to-one.}

\subsection{Implementation and Current State}

The system is implemented in Python 3.11+. The codebase is organized into two modules: one contains the parser, the novelty orchestrator, and the three dimensions; the other, a formalization orchestrator with multi-model abstraction. The verdict tree is identical regardless of which orchestrator formalized the paper.

The project has 153 passed tests and 1 skipped, an evaluation set of 24 theorems, and a dataset of 26 papers withdrawn for duplication with controls matched by category and year.

\section{Instrument Validation}

\label{sec:instrumentos}

Before evaluating the system against real papers (section~\ref{sec:experimentos}), this section validates each instrument separately against known ground truth: D3 against calibration pairs with human judgment, D2 against the evaluation set, and D1 C$_F$ against the coverage of the same set.

\subsection{D3: Calibration Ladder}
\label{ssec:d3cal}

The Jaccard distance was validated on five theorem pairs for which a human judge (the first author) determined the expected relationship between their proofs and signed each label before running the measurement. The pairs cover the spectrum: same proof (\textit{self} control), genuinely different proofs, apparently different proofs that turned out identical after formalization, and unrelated theorems (cross control).

\begin{table}[htbp]
\centering
\small
\setlength{\tabcolsep}{5pt}
\begin{tabular}{llrrrr}
\toprule
Pair & Judge label & Jaccard & Distance & $|\cap|$ & $|\cup|$ \\
\midrule
T08a vs.\ T08a & control: same proof      & 1.0000 & 0.0000 & 9 & 9 \\
T07a vs.\ T07b & \textit{same disguised}    & 0.5000 & 0.5000 & 1 & 2 \\
T08a vs.\ T08b & \textit{genuinely different} & 0.2778 & 0.7222 & 5 & 18 \\
T09a vs.\ T09b & \textit{genuinely different} & 0.0000 & 1.0000 & 0 & 6 \\
T07 vs.\ T08   & control: unrelated   & 0.0000 & 1.0000 & 0 & 10 \\
\bottomrule
\end{tabular}
\caption{D3 calibration over five pairs with prior human labeling. Premise sets are post-filter and deduplicated.}
\label{tab:d3cal}
\end{table}

\textbf{T08 (distance 0.7222).} Two genuinely different proofs of the irrationality of $\sqrt{2}$ yield high distance. The five shared premises are foundational arithmetic lemmas; the remaining ones, exclusive to each side, capture the divergent strategies (divisibility by primes versus 2-adic valuation). The threshold $\theta = 0.5$ correctly classifies the pair as distant proofs.

\textbf{T09 (distance 1.0).} The two proofs of the Gauss sum, induction with \texttt{sum\_range\_succ} versus closed formula with \texttt{Finset.sum\_range\_id}, share no premises after filtering. Each side contributes six premises exclusive to its strategy, and the maximum distance correctly reflects the difference labeled by the judge.

\textbf{T07 (distance 0.50): degenerate point.} The two proofs of the infinitude of primes (Euclid and Euler) fall exactly on the threshold. After formalization, both collapsed to the same invocation of \texttt{Nat.exists\_infinite\_primes}. With only two total premises after filtering and one shared, the Jaccard is fragile: minor changes in filter behavior would invert the verdict. The judge had labeled them as \textit{same disguised}: distinct statements in the LaTeX source, identical in the elaborated proof term. This pair is not informative for calibration.

\textbf{Negative control.} The distance between a number theory theorem and an analysis theorem is 1.0 with empty intersection, confirming that D3 does not produce spurious similarity across distinct domains.

\textbf{What this ladder calibrates.} The five pairs are consistent with human judgment under $\theta = 0.5$. However, three of them (T09 and the two controls) fall at the extremes of the range and do not constrain the choice of threshold: any value in $(0,1)$ classifies them correctly. The only effective bounds come from T08, which requires $\theta < 0.7222$, and from T07, which requires $\theta \geq 0.50$. Since T07 is degenerate (union of two premises, intersection of one, such that a minor change in filter behavior would invert the verdict), the lower bound is not reliable. The admissible interval supported by evidence thus reduces to $\theta < 0.7222$, and the value $\theta = 0.5$ is a design choice within that range, not a calibrated optimum. Narrowing that interval requires pairs whose distance falls in the interior of the range and whose premise sets are large enough for the Jaccard ratio to be stable; constructing them is future work (section~\ref{sec:conclusion}).

\textbf{D3 measures formalizations, not ideas.} The T07 case exposes a limit of the metric that goes beyond the choice of threshold. Mathlib typically contains a single canonical proof of each classical theorem, so that two strategies a mathematician would consider distinct can converge to the same lemma upon being formalized. The consequence is that D3 measures distance between \textit{formalizations}, and the relationship between that distance and the distance between the underlying informal ideas remains unestablished: a distance of 1.0 may be due to the formalizer choosing different lemmas, and a low distance to convergence to the same Mathlib lemma. We report D3 as relative to the formalization over Mathlib v4.29.0. Quantifying how often that convergence occurs requires a systematic experiment across multiple theorems and multiple formalizers, which remains as future work.

\subsection{D2: Triviality Filter}

D2 was evaluated on 24 of the 26 theorems in the evaluation set. Two cases (T20 and T21) were not formalized in this run for reasons unrelated to the instrument. The 24 evaluated cover trivial-by-design cases, classics present in Mathlib, pairs with different proofs, near statements, LLM-generated cases, and deliberate failure cases.

\begin{table}[htbp]
\centering
\small
\setlength{\tabcolsep}{5pt}
\begin{tabular}{llll}
\toprule
Case & Statement & Tactic & Expectation \\
\midrule
T14 & sum of four evens is even        & \texttt{aesop}    & trivial \\
T15 & $2 + 2 = 4$                        & \texttt{decide}   & trivial \\
T16 & $n + 0 = n$                        & \texttt{norm\_num} & trivial \\
T17 & $n \leq n + 1$                     & \texttt{norm\_num} & trivial \\
T19 & LLM-generated about evens       & \texttt{aesop}    & ambiguous \\
T22 & $n$ even $\Rightarrow n+0$ even      & \texttt{norm\_num} & designed for D1 \\
\bottomrule
\end{tabular}
\caption{The six statements that D2 closed with a tactic from \texttt{T\_AUTO}. The first four had a defined triviality expectation; T19 and T22 are discussed in the text.}
\label{tab:d2trivial}
\end{table}

T19 was generated by an LLM instructed to state and prove an original theorem about even numbers; D2 closed it with \texttt{aesop}, which is consistent with the hypothesis that models tend to produce trivial statements, but its expectation in the evaluation set was not binary. T22 is logically equivalent to ``if $n$ is even then $n$ is even''; D2 closed it correctly, but the case was constructed to test syntactic equivalence in D1, not triviality. Neither admits a correct/incorrect classification for D2.

\begin{table}[htbp]
\centering
\small
\setlength{\tabcolsep}{5pt}
\begin{tabular}{p{4.6cm}p{6.4cm}l}
\toprule
Cases & Category & D2 Result \\
\midrule
T01--T06                & classics present in Mathlib            & non-trivial \\
T07a, T08a, T09a        & pairs with different proofs          & non-trivial \\
T10--T13                & near statements (odd primes, AM--GM) & non-trivial \\
T18                     & sum of first $n$ odds $= n^2$ & non-trivial \\
T23                     & connected acyclic graph is a tree         & non-trivial \\
T24                     & coherent sheaves on Noetherian scheme & non-trivial \\
T25                     & $n$ even $\iff 2 \mid n$                  & non-trivial \\
T26                     & sum of $n$ evens is even                 & non-trivial \\
\bottomrule
\end{tabular}
\caption{The 18 statements that D2 correctly classified as non-trivial.}
\label{tab:d2notrivial}
\end{table}

\textbf{Notable cases.} T18 is a control trap: it requires induction and D2 correctly did not close it. T23 was also not closed in this run, unlike previous runs where \texttt{tauto} closed it on a conjunctive definition; the Lean statement effectively used here differs from the one that produced that false positive. T14, in contrast, was closed by \texttt{aesop} in this run, whereas in previous runs with a more complex statement the same tactic exceeded the budget by more than an order of magnitude. Both cases illustrate D2's sensitivity to the concrete statement produced by the formalizer, discussed in section~\ref{sec:limitaciones}.

\subsubsection*{Aggregate}

Over the 24 evaluated theorems, D2 succeeds on 22 (91.7\%). T19 and T22 remain in the denominator but not in the numerator, lacking a binary expectation against which to measure success; over the 22 cases with defined expectation no false positives or false negatives are recorded.

\subsection{D1 C$_F$: Formal Corpus Coverage}

The C$_F$ branch found a match for the 18 non-trivial theorems that reached that stage. The remaining six do not register a match because D1 was never executed on them: they are exactly the six that D2 detected as trivial, on which the tree applied short-circuit. The absence of match in those cases reflects non-execution, not a search failure.

\textbf{This result should not be read as measured coverage.} The C$_F$ match criterion accepts the first result returned by Leandex without applying a similarity threshold, because the current API version does not provide scores (section~\ref{sec:pipeline}). Such a criterion can only return absence of match when the API responds empty, so that 18 matches out of 18 queries does not distinguish real coverage from absence of filtering. The 18 cases correspond to classical theorems that are indeed in Mathlib, and the composition of the evaluation set (dominated by canonical results) makes that outcome expected under any criterion. Establishing the precision of C$_F$ requires a similarity threshold or a manual verification of each match, and neither is available in this version.

The C$_I$ branch was not activated for any theorem. The cause is not the embedding threshold but the entry condition: C$_I$ is only consulted when D2 is negative and C$_F$ found no match, and that combination never occurred. All non-trivial cases had a C$_F$ match, and all those that did not were stopped at D2. The evaluation set, by composition, does not exercise that branch.

\subsection{Summary}

\begin{table}[htbp]
\centering
\small
\setlength{\tabcolsep}{4pt}
\renewcommand{\arraystretch}{1.15}
\begin{tabular}{p{1.9cm}p{2.6cm}p{6.4cm}p{3.4cm}}
\toprule
Instrument & Ground truth & Result & Status \\
\midrule
D3 &
5 pairs with prior human label &
5/5 consistent with the judge under $\theta = 0.5$. Only T08 and T07 constrain the threshold, and T07 is degenerate ($|\cup| = 2$): the evidence supports $\theta < 0.7222$, not a point value. &
Functional; $\theta$ not calibrated \\
\addlinespace
D2 &
24 theorems from the evaluation set &
22/24 successes (91.7\%); T19 and T22 in the denominator without binary expectation. No false positives or negatives over the remaining 22. &
Functional \\
\addlinespace
D1 C$_F$ &
18 non-trivial theorems &
18 matches over 18 queries, but without a similarity threshold the result does not distinguish coverage from absence of filtering. &
Inconclusive \\
\addlinespace
D1 C$_I$ &
--- &
Branch not reached: the entry condition was never met with this set. &
Not evaluated \\
\bottomrule
\end{tabular}
\caption{Validation status of each instrument. Only D2 admits an interpretable success measure with the current evaluation set.}
\label{tab:instrumentos}
\end{table}

\section{Experiments}

\label{sec:experimentos}

This section reports the evaluation of the pipeline on real articles. The ground truth comes from articles withdrawn from arXiv whose withdrawal comment declares duplication of prior results. The evaluation produced both operational measures and the characterization of three obstacles that limit the approach regardless of this implementation.

\subsection{Construction of the Withdrawn Article Set}

The arXiv public API was queried with a search for withdrawn articles in mathematics categories, returning approximately 2,600 results. Over the withdrawal comment text, 23 regular expression patterns were applied, designed to identify withdrawals due to duplication of prior results and exclude withdrawals due to errors, gaps, or administrative issues. The most frequent patterns correspond to formulas such as \textit{result was already known}, \textit{had already been proved by}, \textit{results are not new}, and \textit{subsumed by}.

The filtering yielded 33 candidates, of which 26 proved viable at the time of construction: downloadable LaTeX source and at least one detectable theorem environment. The remaining 7 use formats the parser does not recognize, such as AMS-\TeX{} or abbreviated environment names. The 26 viable ones cover 17 categories, with concentration in combinatorics and algebraic geometry, and a year range from 2001 to 2026. For each, two controls were selected from the same category, with publication year within $\pm 1$, not withdrawn, and with verifiable source; 382 candidates were examined to obtain 52 controls.

\textbf{Ground truth quality.} Of the 26 withdrawn articles, 12 explicitly identify the prior work that duplicates their result; the remaining 14 use generic formulas without specifying the source. Among the 12 that do identify it, only a minority provide an arXiv identifier: the majority cite by author and publication, which requires manual resolution. That asymmetry has consequences analyzed in section~\ref{ssec:cobertura}.

\subsection{Source Availability: A Structural Obstacle}
\label{ssec:fuentes}

During the expansion of the set, a restriction affecting the entire method was discovered: \textbf{arXiv removes the source code of an article upon withdrawal.} The source endpoint returns 404 for withdrawn articles, with no exceptions observed.

This was verified on a random sample of 25 articles withdrawn for duplication drawn from the WithdrarXiv dataset~\cite{withdrarxiv2024}: none retained downloadable source. The hypothesis that articles marked as subsumed but not formally withdrawn would retain their source proved false: on arXiv, the subsumption mark implies withdrawal.

The consequences are twofold. First, no pipeline that starts from LaTeX source can process withdrawn articles, unless the sources were downloaded before withdrawal. Second, and more serious for evaluation, the set we constructed is not reproducible from arXiv: the articles that were viable at the time of construction ceased to be so, and their processable material survives only in local copies. Any future work using withdrawn articles as duplication ground truth must account for this restriction from the design stage, preserving sources at the time of selection.

\subsection{Formalization Model Selection}

Four language models were compared on the \textit{statement-only} formalization task over five withdrawn articles.

\begin{table}[htbp]
\centering
\small
\begin{tabular}{lcl}
\toprule
Model & Success & Predominant failure mode \\
\midrule
Qwen 3.7-max      & 5/5 & --- \\
GLM-5.2           & 3/5 & fixable errors, one timeout \\
DeepSeek V4 Pro   & 0/5 & synthesis errors, placeholder definition \\
DeepSeek V4 Flash & 0/5 & synthesis errors, unexpected tokens \\
\bottomrule
\end{tabular}
\caption{Comparison of formalization backends over five statements. Qwen 3.7-max was the only one that produced mathematically substantive definitions in all five cases, and is the model used in the rest of the experiments.}
\label{tab:modelos}
\end{table}

\textbf{Caveat on this selection.} The five articles in the comparison bank are the same that make up the withdrawn group of the in-depth study. The model was therefore chosen over part of the evaluation set, which introduces an optimistic bias on its performance in that group and offers an alternative explanation for the fidelity asymmetry reported below.

\subsection{In-depth Study}

The in-depth study evaluated 10 articles (5 withdrawn and 5 matched controls) with \textit{statement-only} formalization of the statement, triviality filter, and search in formal and informal corpora. The withdrawn articles were manually selected from the 26 viable ones as clear cases of duplication.

Of the 10 articles, 7 were successfully formalized. Of the remaining 3, one failed due to a compilation error and two due to exceeding the model limit with statements of four levels of nested cases.

\subsubsection{Fidelity Audit}

The first author manually reviewed the 7 successful formalizations, comparing the generated Lean code against the original statement.

\begin{table}[htbp]
\centering
\small
\begin{tabular}{llll}
\toprule
Article & Role & Fidelity & Problem detected \\
\midrule
1609.02090v1   & withdrawn & faithful & --- \\
1207.0631v1    & withdrawn & faithful & --- \\
1212.0196v1    & withdrawn & faithful & --- \\
1004.3381v1    & withdrawn & faithful & --- \\
1101.3720v1    & control  & incorrect & \texttt{sorry} in the central definition \\
0904.1783v3    & control  & approximation & only one direction of an equivalence \\
math/0504586v2 & control  & incorrect & event and measure trivialized \\
\bottomrule
\end{tabular}
\caption{Manual audit of the seven successful formalizations.}
\label{tab:fidelidad}
\end{table}

The fidelity rate is 4/4 among withdrawn articles and 0/3 among controls. Two explanations compete. The first is a complexity bias: the controls, matched by category and year but not by statement difficulty, tend toward longer statements with more nested cases, a hypothesis reinforced by the two controls that failed due to exceeding the model limit. The second, simpler, is the model selection bias noted above: the formalizer was chosen for its performance on these same withdrawn articles and on no controls. The available data do not allow distinguishing between the two.

The experimental consequence is that the three evaluated controls do not contribute evidence on false positives: their formalizations do not represent the original theorems, so that any verdict on them is informative about the formalization and not about the detector.

\subsubsection{Five Failure Modes of Compilation-Based Verification}
\label{ssec:modosfalla}

Automatic formalization introduces a problem that does not exist in manual verification: a Lean file can compile without errors without containing a faithful formalization of the original theorem. During development, five successive failure modes were documented. Each motivated a new guard, and each guard let the next one through.

\begin{enumerate}
\item \textbf{Empty file.} A file with no declarations compiles cleanly. In the first run, five out of five formalizations were empty files that satisfied the original success criterion. The corresponding guard requires the presence of at least one substantive declaration.

\item \textbf{Placeholder definition.} A definition of the form \texttt{def CongruentNumber := True} compiles, contains a substantive declaration, and satisfies the previous guard, but captures no mathematical definition. The response was to reinforce the prompt against definition trivialization.

\item \textbf{Correct definition without the theorem.} After that reinforcement, the same article produced a mathematically correct definition but omitted the theorem statement. The file compiled, the definition was faithful, and the object to be evaluated was absent.

\item \textbf{\texttt{sorry} in an auxiliary definition.} A \texttt{sorry} inside the body of a definition compiles and satisfies any syntactic guard over the declarations: the signature exists, the content is a placeholder. The success criterion rejects files with \texttt{sorry}, but only if the verifier detects it in the right place.

\item \textbf{Only one direction of an equivalence.} A theorem stated as an equivalence in the original article was formalized in only one direction. The file compiles, the definitions are faithful, the formalized direction is correct, and half the statement is missing.
\end{enumerate}

The five steps share a property: compilation verifies coherence, never semantic fidelity. Each guard closes one failure class and lets the next through because the underlying question (whether the generated Lean code means the same as the original statement) is not decidable by syntactic means. Hence human auditing is irreducible: no automatic test can establish that defining a percolation event as the total set is not an acceptable formalization.

This observation has scope beyond AViD. Automatic formalization pipelines routinely report successful compilation rates as a performance measure. The five failure modes show that this metric and semantic fidelity are distinct magnitudes, and that the former does not bound the latter. What fraction of compiling formalizations is semantically unfaithful is an open question requiring systematic auditing with a blind protocol, and one that this work does not answer.

\subsection{Duplication Detection}

\subsubsection{Two Corrected Defects}

Two pipeline defects were identified by running the evaluation, and both produced duplication false positives.

\textbf{Absence of temporal ordering.} Article 1207.0631v1, published in 2012, received a \texttt{KNOWN\_IN\_LITERATURE} verdict based on a 2018 candidate that the judge classified as equivalent. A later work cannot be what an earlier work duplicates: the verdict was logically impossible. The incorporation of the temporal filter (section~\ref{ssec:d1}) discards the three candidates later than the article's date and the verdict correctly changes to \texttt{STATEMENT\_NOVELTY}.

\textbf{Self-article retrieval.} The queried indices contain the evaluated articles. Without explicit exclusion, three of the seven articles retrieved themselves as candidates and the judge classified them as equivalent or specializations, producing non-novelty verdicts without content. With exclusion, the three verdicts are corrected and no self-matches are observed.

Both defects are generic: they affect any system that verifies novelty by retrieval over indices that contain the evaluated material. The second, moreover, had been corrected in one branch of the system and remained active in another, illustrating that a point fix does not propagate on its own.

\subsubsection{Results}

With both corrections and queries constructed from the statement, the pipeline detected a single match on a control article (whose result appears in an earlier indexed work) and none on the withdrawn articles with faithful formalization. That match falls on a control whose formalization was only an approximation (one direction of an equivalence), so that, as noted above, it does not constitute clean evidence about the detector: its verdict informs about the formalization as much as about the search.

The absence of detections among withdrawn articles is not a detector failure. The duplicators of those articles (results by Hardy and Littlewood, Monsky, and Gyárfás and Lehel) predate the electronic preprint era. Section~\ref{ssec:cobertura} shows that this fact, and not the metric or the judge, determines the result.

\subsubsection{Similarity and Result Identity}

One case illustrates the distance between retrieving a similar work and retrieving the duplicated work. Article 1609.02090v1 contains two independent results: one that the article itself explicitly attributes to a prior work, and another (the reason for withdrawal) that had been proved by Hardy and Littlewood around 1928. The search retrieved the first, that is, the work that duplicates a result the author never claimed as their own, and not the second. The match is correct and at the same time irrelevant to the question posed.

\subsection{Coverage of the Retrieval Corpora}
\label{ssec:cobertura}

The above results suggest that the limiting factor lies not in the pipeline but in what the indices contain. To verify this, coverage was measured directly on the known duplicators, without pipeline intervention: the question is whether the duplicated work is indexed, not whether the system finds it.

\textbf{Retrievability of the duplicators.} Of the duplicators identified in the withdrawal comments, the majority correspond to works before 1991: classical results published in journals, not preprints. Only a minority have an arXiv identifier. This distribution is not accidental: when a result is re-proved without noticing, the original work tends to be an established and old result, precisely the type of material that theorem indices cover worst.

To rule out that the absence was an artifact of the query method, each of the twelve duplicators explicitly identified was searched by its exact name in both corpora (TheoremSearch and Matlas), without obtaining any match. The exact-name search, more permissive than the statement-based retrieval used by the pipeline, confirms that those works are not in the reachable corpus: the absence is one of coverage, not of method.

\textbf{Consequence.} A duplicator absent from the index imposes a recall ceiling of zero, regardless of the quality of the similarity metric, the embedding threshold, or the judge. No pipeline improvement modifies that ceiling. The improvement path is corpus expansion, and the incorporation of a peer-reviewed journal index with coverage since 1826 was not sufficient for the evaluated cases. This limitation is not specific to AViD: any system that verifies novelty by retrieval faces it with the currently available infrastructure.

\subsection{A Semantic Similarity Baseline}

As a reference, it was evaluated whether a semantic similarity metric applied directly to the text of statements, without formalization or decision tree, allows distinguishing withdrawn articles from controls. Over 37 evaluable articles, a Mann--Whitney test on the score distributions yields no significant difference ($p = 0.854$).

This result should not be read as evidence of absence of effect. The sample size is small and uneven, the exclusion of articles was not random (controls had approximately twice the probability of being excluded due to absence of extractable statement), and statistical power is not reported. What the result does indicate is that textual similarity between statements, taken in isolation, is not a useful discriminator of duplication in this set.

A previous version of this same baseline yielded an apparently positive result that proved artifactual: without self-article exclusion, the majority of strong matches were self-matches. It is documented here as a warning for works that evaluate over indices that contain the evaluated material.

\subsection{Summary}

\begin{table}[htbp]
\centering
\small
\setlength{\tabcolsep}{4pt}
\renewcommand{\arraystretch}{1.15}
\begin{tabular}{p{3.6cm}p{1.6cm}p{8.6cm}}
\toprule
Evaluation & $n$ & Result \\
\midrule
Backend selection & 5 $\times$ 4 & Qwen 3.7-max 5/5; both DeepSeek models 0/5. Selection contaminated by overlap with the evaluation set. \\
\addlinespace
Formalization & 10 & 7 formalized. Fidelity 4/4 on withdrawn, 0/3 on controls; controls do not contribute evidence on false positives. \\
\addlinespace
Failure modes & --- & Five successive modes in which successful compilation does not imply semantic fidelity. \\
\addlinespace
Duplication detection & 7 & Two defects corrected (temporal order, self-match). No detections on withdrawn articles with pre-1991 duplicator. \\
\addlinespace
Corpus coverage & --- & The content of the indices, and not the metric, determines the recall ceiling. \\
\addlinespace
Source availability & 25 & No withdrawn article source remains available on arXiv. \\
\addlinespace
Semantic baseline & 37 & No significant difference ($p = 0.854$); small and biased sample. \\
\bottomrule
\end{tabular}
\caption{Evaluation summary. The last three rows correspond to structural obstacles, not to pipeline performance measurements.}
\label{tab:resumen}
\end{table}

\section{Conclusion}

\label{sec:conclusion}

\subsection{What Was Built}

AViD Journal traverses the complete path from a LaTeX article to a novelty verdict: parser of mathematical environments with topological ordering of dependencies, formalization in Lean~4 with abstraction over eight backends, three reference corpora (Mathlib via Leandex, TheoremSearch, and Matlas), and an eight-verdict decision tree over three dimensions evaluated in increasing cost order.

\subsection{What Was Measured}

D2 succeeded on 22 of 24 theorems in the evaluation set (91.7\%), with no false positives or false negatives over the cases with defined expectation. D3 proved consistent with human judgment in the five calibration pairs, although only two of them constrain the threshold. D1 C$_F$ returned a match in the 18 queries that reached that stage, but without a similarity threshold that result does not distinguish real coverage from absence of filtering. On real articles, the pipeline detects duplication when the duplicated work is indexed in the corpus, and does not detect it when it is not.

The comparison of four formalization models quantifies the relative viability of different backends and motivated the choice of the one used in the experiments. Additionally documented is a methodological defect that affects any evaluation over theorem indices: without explicit self-article exclusion, a paper retrieves itself as a candidate and the judge classifies it as equivalent, producing duplication false positives.

\subsection{What Was Learned}

The evaluation produced a result more informative than any pipeline performance measure: the identification of three obstacles that are not deficiencies of this implementation but properties of the problem with the infrastructure available today.

The first is that compilation verifies coherence, never semantic fidelity: a Lean file can compile without errors and not represent the original statement, and no syntactic guard closes that gap (section~\ref{sec:experimentos}). The second is that the coverage of theorem indices, and not the similarity metric or the judge, imposes the recall ceiling: a duplicator absent from the corpus cannot be found by any pipeline, however good. The third is that arXiv removes the source code of articles upon withdrawal, which restricts what can be processed and compromises the reproducibility of any benchmark built upon them: the set of articles we assembled is replicable in its composition (identifiers and withdrawal comments) but not in its processable material, which must be preserved locally before withdrawal.

The three are transferable: any system that verifies novelty by retrieval and formalization will encounter them.

\subsection{Limitations and Future Work}
\label{sec:limitaciones}

The remaining limitations of this version define, for the most part, the work that follows.

\textbf{Scope of the metric.} AViD evaluates each statement separately, so that an article whose novelty lies in the combination of known results would not be detected; an aggregation layer is future work. D3 operates only over \texttt{Prop} and assumes proof irrelevance: extending it to \texttt{Type}, where the appropriate notion would be homotopic, is a long-term goal. More fundamentally, the framework measures novelty over statements and proofs frozen in a corpus and does not capture the conceptual novelty of a definition that reformulates a problem or of a method transferable to other domains~\cite{lakatos1976}; and proof identity is undecidable in general~\cite{dosen2003}, so that Jaccard over premises is an engineering approximation, with false positives when equivalent proofs use superficially distinct premises and false negatives when distinct strategies share core lemmas.

\textbf{D3 calibration.} The threshold $\theta = 0.5$ is a design choice. The calibration ladder is consistent with human judgment in the five pairs, but only two constrain the threshold and one of them is degenerate, so that the evidence bounds the admissible interval to $\theta < 0.7222$ without fixing a value. Calibrating requires a corpus of proof pairs for the same theorem with prior human labeling, selected so that their distances fall in the interior of the range and their premise sets are large enough for the ratio to be stable; such a corpus would also allow weighting premises by IDF~\cite{huch2022} instead of counting them equally, replacing premise sets with complete dependency graphs, and reporting precision and recall of D3 on the proof identity axis. This is the first thing we will do in the next version.

\textbf{Formalizer dependence.} AViD evaluates the formalized proof, not the informal one. A maximum distance may be due to the formalizer choosing different lemmas, and a low distance to convergence to the same Mathlib lemma (section~\ref{ssec:d3cal}). The same effect operates on D2, whose result depends on the triple (formalized statement, tactic, budget) and not on the theorem. Quantifying how often self-formalization preserves or collapses the distinction between strategies requires a systematic experiment across multiple theorems and multiple models, with human auditing of the fidelity of each formalization; it is the line of work we consider most promising. Running several formalizers in parallel would also reduce dependence on a single provider.

\textbf{D1 precision.} Statement comparison is syntactic: incorporating \texttt{isDefEq} would allow recognizing definitional equivalences that currently produce false negatives. The informal branch also depends on being able to formalize third-party proofs, a capability that had 0\% success in the proof of concept; without that link D3 cannot be applied to informal corpus matches.
Two extensions point to accelerated publication scenarios like the one described in the introduction. The first is of temporal granularity: the date that the filter derives from the arXiv identifier resolves at the month level, sufficient to discard candidates later by years but not to establish priority between works appearing days apart; replacing it with the submission date provided by the API would resolve that case. The second is of scope: the pipeline formalizes statements and compiles the result, so that the infrastructure for evaluating the validity of a counterexample (a computationally much cheaper task than verifying a proof) is already present, although the current decision tree only issues novelty verdicts. Extending it in that direction would require first measuring with what fidelity the formalizer translates counterexample statements, and therefore faces the same obstacle documented in section~\ref{ssec:modosfalla}.

\textbf{Evaluation scale.} The 24 theorems in the evaluation set were constructed by the first author, and the in-depth study operated on articles selected from those that proved parseable and formalizable, introducing a technical accessibility bias whose effect on duplicability we do not know. Three directions expand that scale: evaluating D2 against a corpus of independently labeled conjectures, which would provide a measure with external ground truth; running the complete pipeline over Mathlib, which offers a formal corpus of magnitude far greater than the current set; and building a benchmark over self-formalized arXiv articles, where the source remains available. As for the duplication ground truth, the withdrawal comment is an imperfect proxy: of the viable withdrawn articles, fewer than half explicitly identify the work that duplicates them, and among those that do the majority cite it by author and publication rather than by identifier, which requires manual resolution with its own margin of error.

\textbf{Engineering.} Premise extraction requires loading full Mathlib, so that D3 is offered as on-demand analysis and not within the interactive flow; imports of specific modules do not work on temporary files, which forces loading the entire library on each D2 invocation.

\subsection{Closing}

AViD Journal does not solve the problem of automatic novelty verification. It poses it operationally, implements it end to end, and documents precisely where it fails. The system operates over the subset of theorems expressible in the \texttt{Prop} fragment of Lean~4, with statement parseable and formalizable by some available model; within that perimeter the decision tree produces verifiable verdicts, and outside it, the system is explicitly incomplete.

The thesis of the work is that the question of whether novelty can be automatically arbitrated is not answered with a metric design or an architecture argument, but by building the system, running it against real ground truth, and reporting what worked and what did not. The three identified obstacles suggest that the bottleneck is not where metric design invites one to look for it.


\begin{thebibliography}{99}

\bibitem{abouzaid2026}
M.\textasciitilde{}Abouzaid \emph{et al.}, ``First Proof,''
arXiv:2602.05192, 2026.

\bibitem{firstproof2b}
M.\textasciitilde{}Abouzaid, N.\textasciitilde{}Srivastava, R.\textasciitilde{}Ward, and L.\textasciitilde{}Williams,
``First Proof: Second Batch,'' First Proof Foundation, June 2026.
\url{https://1stproof.org/assets/docs/report.pdf}

\bibitem{alpoge2026}
L.\textasciitilde{}Alp\"oge, ``Counterexample to the Jacobian conjecture in dimension~3,''
announcement on X, July 19, 2026,
\url{https://x.com/__alpoge__/status/2079028340955197566}.
Digest: T.\textasciitilde{}Tao, ``A digestion of the Jacobian conjecture counterexample,''
2026, \url{https://terrytao.wordpress.com/2026/07/21/}.
Independent formal verification: \emph{Jacobian\_Counterexample}, Archive of Formal Proofs, 2026.

\bibitem{tao2025}
T.\textasciitilde{}Tao, ``Machine-Assisted Proof,''
\emph{Notices of the AMS}, 2025.

\bibitem{kasaura2025}
K.\textasciitilde{}Kasaura \emph{et al.}, ``Discovering New Theorems via LLMs with In-Context Proof Learning in Lean,''
arXiv:2509.14274, 2025.

\bibitem{theoremsearch2026}
L.~Alexander, E.~Leonen, S.~Szeto, A.~Remizov, I.~Tejeda, G.~Inchiostro, and V.~Ilin,
``Semantic Search over 9 Million Mathematical Theorems,''
arXiv:2602.05216, 2026.

\bibitem{theoremgraph2026}
S.~Kurgan, E.~Wang, E.~Leonen, S.~Szeto, L.~Alexander, A.~Remizov, J.~Alper,
G.~Inchiostro, and V.~Ilin, ``TheoremGraph: Bridging Formal and Informal Mathematics,''
arXiv:2606.25363, 2026.

\bibitem{compose2026}
D.~Busbib and M.~Werman, ``COMPOSE: Composing Future Theorems from Citations and Formal Structure,''
arXiv:2605.30333, 2026.

\bibitem{leanconjecturer2025}
N.\textasciitilde{}Onda \emph{et al.}, ``LeanConjecturer: Filtering Novelty in LLM-Generated Conjectures,''
arXiv:2506.22005, 2025.

\bibitem{matlas2026}
H.\textasciitilde{}Ju \emph{et al.}, ``Matlas: A Semantic Search Engine for Mathematics,''
arXiv:2604.17484, 2026.

\bibitem{yoo2025}
H.\textasciitilde{}Yoo, ``The Axiom-Based Atlas: A Structural Mapping of Theorems via Foundational Proof Vectors,''
arXiv:2504.00063, 2025.

\bibitem{magnushammer2024}
M.\textasciitilde{}Miku{\l}a, S.\textasciitilde{}Antoniak, S.\textasciitilde{}Tworkowski \emph{et al.}, ``Magnushammer: A Transformer-Based Approach to Premise Selection,''
arXiv:2303.04488, \emph{ICLR}, 2024.

\bibitem{piotrowski2023}
B.\textasciitilde{}Piotrowski, R.\textasciitilde{}Fern\'andez Mir, and E.\textasciitilde{}Ayers, ``Machine-Learned Premise Selection for Lean,''
arXiv:2304.00994, \emph{TABLEAUX}, 2023.

\bibitem{huch2022}
F.\textasciitilde{}Huch, ``Structure in Theorem Proving: Analyzing and Improving the Isabelle Archive of Formal Proofs,''
arXiv:2209.13305, 2022.

\bibitem{mining2024}
J.~Chuharski, E.~Rojas Collins, and M.~Meringolo,
``Mining Math Conjectures from LLMs: A Pruning Approach,''
arXiv:2412.16177, 2024.

\bibitem{synthetic2024}
``Synthetic Theorem Generation in Lean,''
OpenReview EeDSMy5Ruj, 2024.

\bibitem{survey2026}
H.~Ju and B.~Dong, ``AI for Mathematics: Progress, Challenges, and Prospects,''
arXiv:2601.13209, 2026.

\bibitem{lakatos1976}
I.\textasciitilde{}Lakatos, \emph{Proofs and Refutations}, Cambridge University Press, 1976.

\bibitem{dosen2003}
K.\textasciitilde{}Dosen, ``Identity of Proofs,''
\emph{Bulletin of Symbolic Logic}, vol.\textasciitilde{}9, no.\textasciitilde{}4, 2003.

\bibitem{network2026}
X.~Li, N.~Peng, S.~Severini, and P.~Shafto, ``The Network Structure of Mathlib,''
arXiv:2604.24797, 2026.

\bibitem{leandex2026}
Project Numina, ``Numina-Lean-Agent: An Open and General Agentic Reasoning
System for Formal Mathematics,'' arXiv:2601.14027, 2026.
Leandex: \url{https://leandex.projectnumina.ai}.

\bibitem{leanfinder2025}
J.~Lu, K.~Emond, K.~Yang, S.~Chaudhuri, W.~Sun, and W.~Chen,
``Lean Finder: Semantic Search for Mathlib That Understands User Intents,''
arXiv:2510.15940, 2025; ICLR 2026.

\bibitem{kuhlwein2013}
D.~K\"uhlwein, J.~C.~Blanchette, C.~Kaliszyk, and J.~Urban,
``MaSh: Machine Learning for Sledgehammer,''
\emph{Interactive Theorem Proving (ITP 2013)}, LNCS 7998, pp.~35--50, Springer, 2013.

\bibitem{kaliszyk2014}
C.~Kaliszyk and J.~Urban, ``Learning-Assisted Automated Reasoning with Flyspeck,''
\emph{Journal of Automated Reasoning}, vol.~53, no.~2, pp.~173--213, 2014.


\bibitem{withdrarxiv2024}
R.~Rao \emph{et al.}, ``WithdrarXiv: A Large-Scale Dataset of Withdrawn Papers,''
arXiv:2412.03775, 2024.

\end{thebibliography}
\end{document}